\documentclass{article}

\usepackage[preprint]{neurips_2026}

\usepackage[utf8]{inputenc}
\usepackage[T1]{fontenc}
\usepackage{hyperref}
\usepackage{url}
\usepackage{booktabs}
\usepackage{amsfonts}
\usepackage{amsmath}
\usepackage{amssymb}
\usepackage{nicefrac}
\usepackage{microtype}
\usepackage{xcolor}
\usepackage{graphicx}
\usepackage{tikz}
\usepackage{float}
\usetikzlibrary{shapes.geometric, arrows.meta, positioning, calc}

\title{Do Deep Stereo Models Learn Shape Priors?}

\author{%
  \begin{tabular}{ccc}
    Linbo Tang & Aneel Damaraju & Todd Zickler \\
    Harvard University & Harvard University & Harvard University \\
    \small\texttt{linbotang@fas.harvard.edu} &
    \small\texttt{aneeldamaraju@g.harvard.edu} &
    \small\texttt{zickler@seas.harvard.edu}
  \end{tabular}
}

\newcommand{\boldstart}[1]{\vspace{0.08in}\noindent{\bf #1}}

\begin{document}

\maketitle

\begin{abstract}
Localizing object boundaries is important for embodied visual systems, since small boundary errors can change the inferred geometry of physical surfaces. We study whether deep stereo models produce boundaries consistent with simple shape priors when local image and disparity cues are weak. We create a synthetic stereo dataset consisting of objects drawn from a highly constrained family of shapes, with randomized depth, slant, color and texture. Despite the simplicity of this shape space, we find that several state-of-the-art stereo models retain systematic boundary errors, even when finetuned or trained from scratch on the dataset. We then introduce a simple generative stereo model that jointly predicts disparity and a distance field representing object boundaries. 
The model improves boundary localization and produces contours that adhere more closely to the underlying shape space. 
These results expose limitations in the boundary predictions of current stereo architectures when local evidence is weak, and motivate further study of whether generative models can better combine regional grouping evidence with global geometric structure.
\end{abstract}

\section{Introduction}
\label{sec:introduction}

Localizing object boundaries is important for embodied visual systems that must interact with the physical geometry of their surroundings. Errors near object boundaries can have consequences that are disproportionate to their contribution to average error, because a small displacement changes where one surface ends and another begins. In order to reliably localize boundaries, a vision system must learn to group regions of uniform color, texture and disparity; and it must learn to combine these with shape priors that enable completing boundaries that are ambiguous, such as in cases of partial camouflage.

Deep stereo models are attractive for this because they promise to achieve it from being trained end-to-end with a simple disparity loss. Given enough training data, these models ought to be able to learn to group within-object regions and to localize crisp disparity boundaries between objects with very high precision. 

\begin{figure}[t]
    \centering
    \includegraphics[width=1.0\linewidth]{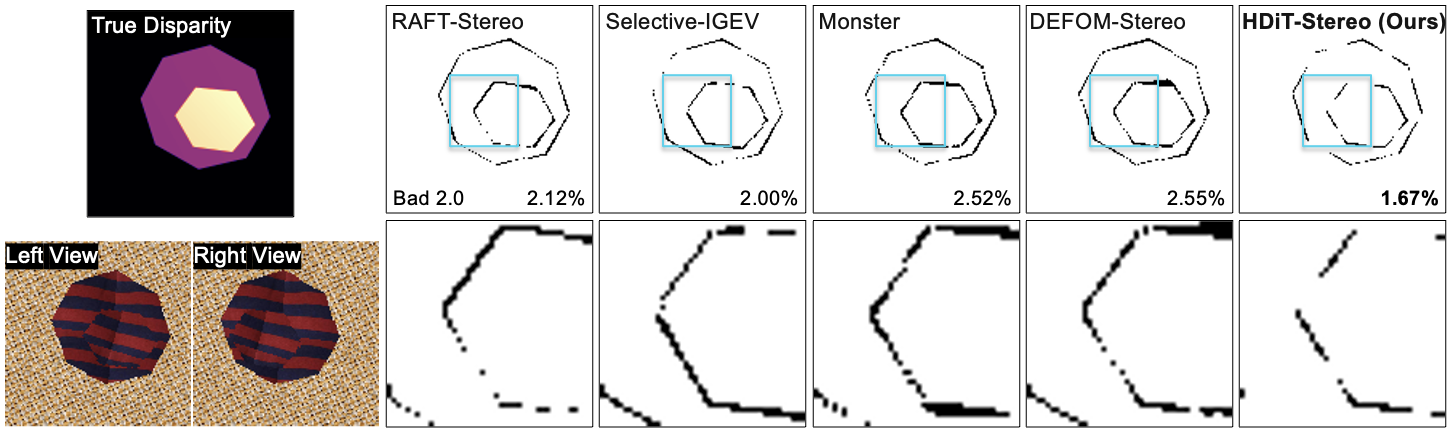}
    \caption{\textbf{Boundary errors when texture and disparity cues are weak.} Left panel: input stereo images and true disparity. Right panel: locations where each model's predicted disparity error is greater than two pixels, along with overall Bad~2.0 percentage. Despite being finetuned or trained from scratch on our heavily constrained shape distribution, several deep stereo models retain substantial errors near object boundaries. 
    Our diffusion-based HDiT-Stereo reduces errors near boundaries, particularly where local evidence is weak.
    }
    \label{fig:teaser_results}
\end{figure}

In this paper, we devise a controlled test of this ability using synthetic stereo images with heavily constrained shapes (regular polygons) rendered using randomized depths, slant angles, colors and textures. Because the space of possible shapes is so constrained, we expect high-capacity stereo models trained on this distribution to be able to internalize its shape statistics sufficiently well to precisely complete crisp boundaries, even when color, texture and disparity cues are weak or partially missing. 
Instead, we find that a variety of state-of-the-art deep stereo models produce distorted boundaries, even when finetuned or re-trained from scratch on the dataset, with errors concentrating where local evidence is weak (Figs.~\ref{fig:teaser_results} and~\ref{fig:shape_prior_qualitative}). These errors are consistent with the models failing to fully exploit the available shape regularities, although boundary accuracy alone cannot distinguish this explanation from improved local contour prediction. 

This motivates us to ask whether we can improve boundary recovery in these settings using a generative formulation, where the model learns a distribution over plausible scene geometry instead of deterministically mapping stereo images to disparity.
We introduce a straightforward diffusion-based architecture that jointly generates disparity and a distance field representing object boundaries. The architecture is deliberately simple, but it measurably improves boundary accuracy and produces recovered contours that adhere more closely to the constrained shape space of our dataset. 

The generative formulation also provides a natural representation of uncertainty when the scene geometry is not completely determined by the images. In particular, when we construct scenes for which two figure–ground interpretations are genuinely consistent with the stereo images, the model produces distinct samples corresponding to the two valid interpretations rather than collapsing them into an invalid averaged contour.

Together, our results suggest that more work needs to be done to achieve accurate physical perception, and that generative models of scene geometry are worth considering as a means of 
representing both local boundary evidence and global geometric structure.

\section{Related work}
\label{sec:related_work}

\boldstart{Stereo and boundaries.} Modern deep stereo models typically combine explicit correspondence representations with learned spatial regularization and iterative refinement, often augmented by monocular grouping priors and long-range spatial reasoning~\citep{lipson2021raftstereo, wang2024selective, cheng2025monster, jiang2025defom,wen2025foundationstereo}. They have achieved increasingly strong disparity accuracy and substantial cross-domain generalization. Although these models are generally trained to predict disparity rather than explicitly represent object boundaries or shape, their learned representations should in principle capture statistical regularities in scene geometry, including smooth disparity within objects and abrupt changes at their boundaries. Our work asks how effectively they exploit these regularities when the space of possible object shapes is deliberately made simple and highly constrained.

Depth boundaries pose a particular challenge for stereo because the local evidence that identifies them can be weak or missing. A model can achieve low average disparity error while shifting, rounding, or blurring the boundaries that determine the recovered scene geometry. Prior work has addressed this through explicit reasoning about occlusions and their boundaries, specialized cost aggregation for textureless regions and disparity discontinuities, and losses designed to better preserve boundary structure~\citep{wang2019local,liu2024goat,xu2020aanet,xu2024adaptive}. 
Our focus is complementary: we ask how accurately learned stereo models localize and complete boundaries in a setting where global shape regularities could compensate for weak local evidence.

\boldstart{Shape priors and perceptual organization.} The ability to infer object shape from incomplete or ambiguous local evidence has a long history in both human and computer vision. Humans readily group spatially separated edge elements into coherent contours and complete object boundaries across occlusion or camouflage, suggesting that local appearance cues are combined with more global constraints on shape and contour organization~\citep{palmer1999vision}. Deep visual models, however, do not necessarily acquire the same biases from standard training objectives. Image classification networks have been shown to rely disproportionately on local texture rather than global shape~\citep{geirhos_imagenet-trained_2018, baker_deep_2018}, and more recent work on contour integration suggests that human-like use of global contour structure emerges only in very large vision models~\citep{lonnqvist_contour_2025}. Our work asks whether similar limitations arise in stereo, where global shape is not itself the prediction target but could, in principle, help resolve uncertain local evidence.

\boldstart{Generative models for geometric prediction.} Generative models provide an alternative to deterministic regression by learning a distribution over plausible outputs conditioned on visual input. Diffusion models have been applied to monocular depth, optical flow, and dense correspondence, demonstrating both competitive geometric estimation and the ability to represent uncertainty or multiple plausible solutions~\citep{saxena2023surprising,nam2024diffusion,ke2024repurposing,zheng2025diffuvolume}. More broadly, recent work suggests that generative image models can provide useful priors for downstream visual understanding~\citep{tang2023emergent, gabeur2026visionbanana}. Our work explores these ideas for stereo, using diffusion not primarily to improve average disparity accuracy, but to learn a distribution over plausible scene shapes and to represent distinct interpretations when boundary geometry is ambiguous.

\section{Task and dataset}
\label{sec:problem_setup}

\label{sec:setup}

\begin{figure}[t]
    \centering
    \includegraphics[width=0.9\linewidth]{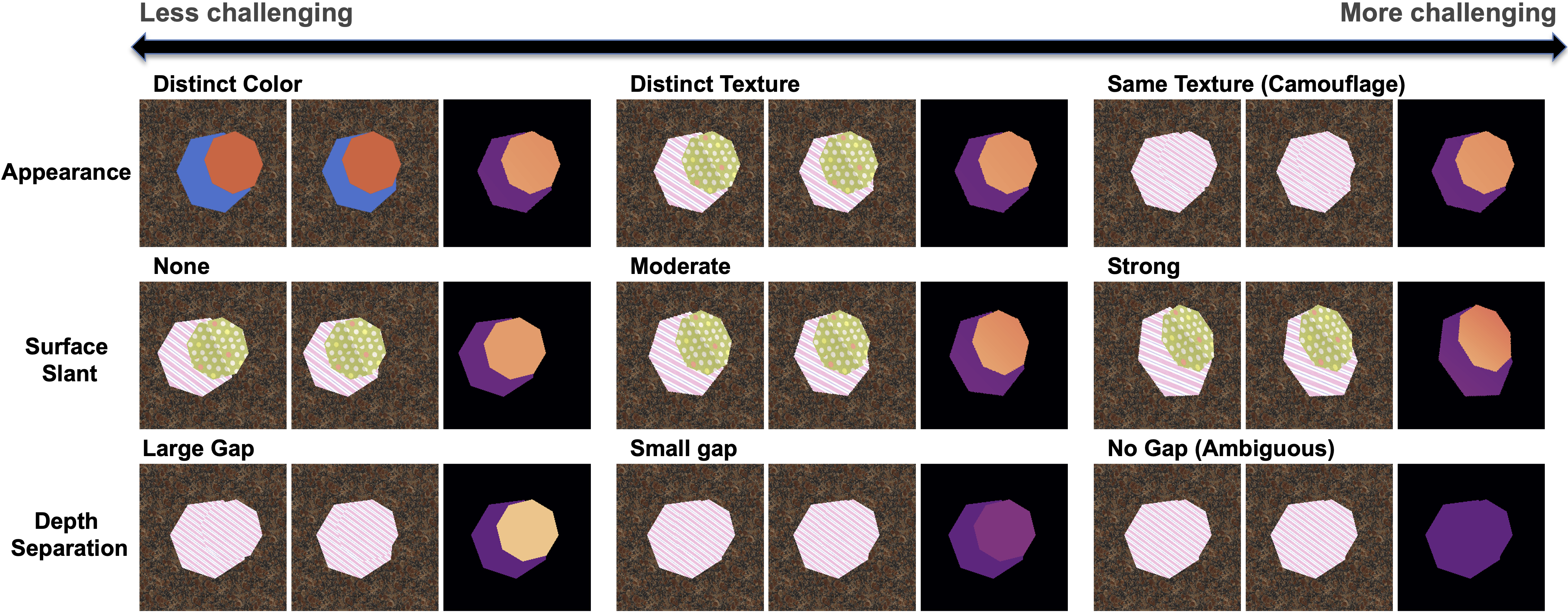}
    \caption{\textbf{Dataset examples.} Shapes are slanted planar regular polygons up to degree eight with randomized appearance. Difficulty is controlled by similarity in object appearance, object slant, and object depth. The extreme case (bottom right) provides no local evidence for much of the objects' boundaries, requiring nonlocal shape information for successful grouping.}
    \label{fig:data_example}
\end{figure}

Given a rectified stereo pair $I_L,I_R \in \mathbb{R}^{3\times H\times W}$, we aim to estimate dense disparity $d\in \mathbb{R}^{H\times W}$ on the left-image grid. A scene point visible in both views at left coordinate $(x,y)$ appears at $(x-d(x,y),y)$ in the right view, and larger disparity indicates smaller depth. Rectification restricts the correspondence search to within each horizontal scanline. 

Our focus is boundary localization and shape recovery when local evidence is weak. Weaknesses occur when neighboring surfaces have similar color or texture, so are partially camouflaged; when textureless regions provide little evidence for identifying left-right correspondences; and when background pixels near occluding contours are visible in only one view, with no valid left-right match. 
Our aim is to assess boundary recovery in conditions where local evidence is weak and learned global shape regularities could be useful.

\label{sec:data}

To create a controlled setting in which such regularities are simple and highly learnable, we construct a procedural dataset of stereo images and true disparity maps using a heavily constrained set of object shapes: regular planar polygons with degree up to eight. We randomize the objects' depths, slants, colors and textures as depicted in Fig.~\ref{fig:data_example}. This approach allows exact rasterization of images and disparity near boundaries, avoiding the errors of sensor-derived data.

Each scene contains one or more objects drawn from the shape library. Objects appear with uniform colors, procedural patterns, or textures from the Describable Textures Dataset~\citep{cimpoi2014describing}. Backgrounds have zero disparity and are also textured or uniformly colored. As shown in Fig.~\ref{fig:data_example}, some scenes are more challenging than others, depending on the distinctiveness of the objects' appearance, their 3D slants, and the size of the depth separation between them. In the most extreme case (bottom right), there are no local appearance or disparity cues for discerning large portions of the objects' boundaries, so in order to succeed in identifying the true boundaries everywhere, a model must rely on nonlocal information about contour continuation and object shape to perform figural scission~\citep{kellman1991theory}, which is known to be ambiguous. (We say more about this in Section~\ref{sec:bimodality}.) 

Associated with each stereo pair, the dataset includes ground truth disparity maps, visible segmentation masks, amodal segmentation masks~\citep{li2016amodal,zhu2017semantic}, and boundary maps represented as unsigned distance fields. (For target continuous boundary points $\mathcal{B}\subset\mathbb{R}^2$ and discrete image grid $\Omega\subset\mathbb{Z}^2$ the rasterized unsigned distance field is $u(\mathbf{p})=\min_{\mathbf{q}\in\mathcal{B}}\lVert\mathbf{p}-\mathbf{q}\rVert_2$ for $\mathbf{p}\in\Omega$.) The dataset includes all of these defined on both a cyclopean-image grid and the left-image grid; we only use the left-image grids in this paper. The dataset generation code will be released alongside the full training and inference pipeline at \url{https://github.com/LINBOTANG123/HDIT-Stereo}.


\section{A simple stereo diffusion model}
\label{sec:model}

\begin{figure}[!htbp]
    \centering
    \includegraphics[width=0.8\linewidth]{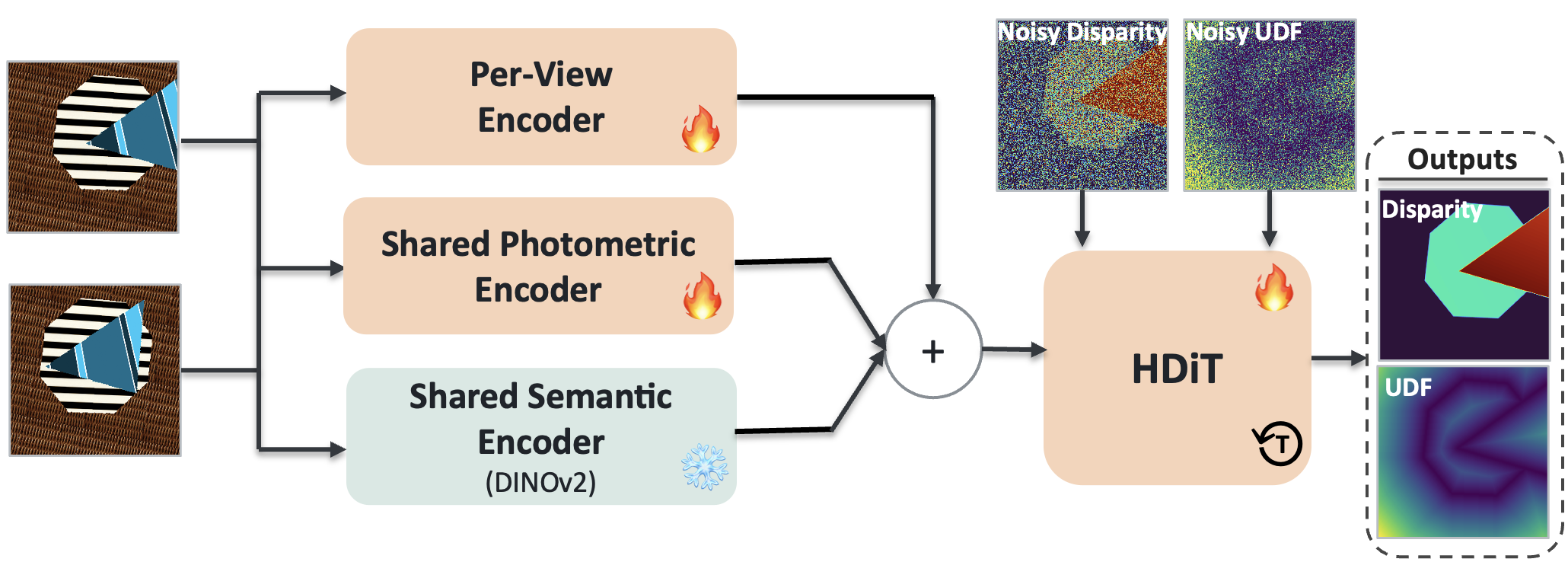}
    \caption{\textbf{Overview of HDiT-Stereo.}
    Per-view image features and correspondence features from a trainable photometric encoder and a frozen DINOv2 semantic encoder condition an HDiT denoiser. From noise the model jointly generates disparity and an unsigned boundary distance field through iterative denoising. Flame and snowflake symbols indicate trainable and frozen components.}
    \label{fig:model_pipeline}
\end{figure}

We contrast deterministic stereo models that produce per-pixel disparity estimates with a generative framework that models a conditional distribution over scene geometry from which we can sample.
We adapt the HDiT diffusion model~\citep{crowson2024hourglass} to stereo and introduce \mbox{HDiT-Stereo}, a generative model for stereo disparity. Fig.~\ref{fig:model_pipeline} shows the  pipeline.

HDiT-Stereo models the conditional distribution $p_\theta\!\left(z \mid c(I_L,I_R)\right)$ over a scene representation $z$,
where $c(I_L,I_R)$ denotes the conditioning features computed from the stereo pair. As shown in Fig.~\ref{fig:model_pipeline}, conditioning features include per-view image features and explicit correspondence features to capture local matching evidence in addition to broader visual context. The photometric and semantic encoders each include feature extraction and left--right correlation over a shared set of candidate disparities. The photometric encoder uses a small CNN trained jointly with the model, while the semantic encoder uses a frozen DINOv2 ViT~\citep{oquab2024dinov2} to provide complementary contextual features.

We obtain a prediction by sampling $z \sim p_\theta\!\left(z \mid c(I_L,I_R)\right)$. HDiT-Stereo jointly generates two normalized fields on the left-image grid, $z=(\tilde{d},\tilde{u})$: disparity $\tilde{d}$ and the unsigned distance $\tilde{u}$ to the nearest object boundary. Disparity represents surface geometry but encodes boundaries only implicitly as discontinuities. In contrast, $\tilde{u}$ represents boundary geometry explicitly through its zero level set, providing a direct target for learning contour location and shape. 

When stereo evidence strongly constrains the geometry, the conditional distribution should concentrate around a single interpretation. When the evidence is ambiguous such as the identical-appearance, equal-disparity configurations illustrated in the bottom-right of Fig.~\ref{fig:data_example}, it can instead represent distinct alternatives (Sec.~\ref{sec:bimodality}). 
Variation across samples also enables uncertainty estimates (Sec.~\ref{sec:uncertainty}).



Training follows the EDM formulation~\citep{karras2022edm}. We corrupt the target fields with Gaussian noise at a randomly sampled scale \(\sigma\) and train the network to recover the clean fields. Per-view image features and correspondence features are added to the HDiT tokens. At inference, we generate disparity and boundary fields by iteratively denoising Gaussian noise conditioned on the stereo pair.

\section{Experiments}

\label{sec:experiments}

\label{sec:setup_exp}

We compare against RAFT-Stereo~\citep{lipson2021raftstereo}, Selective-IGEV~\citep{wang2024selective}, MonSter~\citep{cheng2025monster}, and DEFOM-Stereo~\citep{jiang2025defom}. Each baseline is evaluated in three distinct training regimes: zero-shot using the authors' released weights from training on SceneFlow~\citep{mayer2016sceneflow}; fine-tuned weight using our training set; and trained from scratch using our dataset. We also evaluate FoundationStereo~\citep{wen2025foundationstereo} using its released pretrained weights to test its zero-shot generalization to our dataset, since it is designed to transfer across domains without task-specific fine-tuning.

\begin{table*}[!htbp]
\centering
\small
\setlength{\tabcolsep}{4pt}
\caption{\textbf{Contour fidelity and global accuracy.}
Boundary-localization metrics (left columns) are this paper's focus;
global metrics (right columns) are reported for completeness.
Results use 1000 test scenes and variability is reported as $\pm$ IQR/2 across scenes. Beneath each method are its total and \emph{trainable} parameters in millions. Our model is sampled $10$ times per scene, and we report the average metric across the samples (\emph{Mean}) and the oracle \emph{Best-1-of-10}, which for each metric is the best value among the 10 samples, selected using ground truth. In every column, \textbf{bold} marks the best value and \underline{underline} the second best, excluding \emph{Best-1-of-10}. Our model leads the boundary and global Bad-$X$ columns and is mid-field for global EPE.}


\label{tab:main_1000}
\resizebox{\textwidth}{!}{%
\begin{tabular}{llccc|ccc}
\toprule
& & \multicolumn{3}{c|}{\textbf{Boundary \& contour}}
  & \multicolumn{3}{c}{Global} \\
\cmidrule(lr){3-5}\cmidrule(lr){6-8}
\shortstack[l]{Method\\[-1pt]
{\scriptsize Params: total\,/\,\emph{trainable}}}
& Condition
& Bound.\ EPE $\downarrow$
& Chamfer (px) $\downarrow$
& Hausd.\ 99 (px) $\downarrow$
& EPE $\downarrow$
& Bad1 (\%) $\downarrow$
& Bad2 (\%) $\downarrow$
\\
\midrule

RAFT-Stereo & Zero-shot
& \shortstack{17.192\\{\scriptsize$\pm3.70$}}
& \shortstack{6.987\\{\scriptsize$\pm4.63$}}
& \shortstack{37.96\\{\scriptsize$\pm18.89$}}
& \shortstack{19.868\\{\scriptsize$\pm3.68$}}
& \shortstack{71.00\\{\scriptsize$\pm13.42$}}
& \shortstack{48.43\\{\scriptsize$\pm34.55$}}
\\
{\scriptsize 11.1\,M\,/\,\textit{11.1}\,M} & Train-from-scratch
& \shortstack{\underline{0.792}\\{\scriptsize$\pm0.23$}}
& \shortstack{0.345\\{\scriptsize$\pm0.05$}}
& \shortstack{1.37\\{\scriptsize$\pm0.21$}}
& \shortstack{\textbf{0.118}\\{\scriptsize$\pm0.02$}}
& \shortstack{\underline{2.62}\\{\scriptsize$\pm0.39$}}
& \shortstack{\underline{1.71}\\{\scriptsize$\pm0.33$}}
\\
& Fine-tuned
& \shortstack{0.867\\{\scriptsize$\pm0.23$}}
& \shortstack{0.382\\{\scriptsize$\pm0.06$}}
& \shortstack{1.74\\{\scriptsize$\pm0.50$}}
& \shortstack{0.142\\{\scriptsize$\pm0.03$}}
& \shortstack{2.86\\{\scriptsize$\pm0.41$}}
& \shortstack{1.86\\{\scriptsize$\pm0.37$}}
\\
\midrule

Selective-IGEV & Zero-shot
& \shortstack{5.892\\{\scriptsize$\pm2.25$}}
& \shortstack{6.633\\{\scriptsize$\pm2.57$}}
& \shortstack{34.78\\{\scriptsize$\pm12.03$}}
& \shortstack{5.510\\{\scriptsize$\pm2.65$}}
& \shortstack{39.62\\{\scriptsize$\pm29.22$}}
& \shortstack{33.59\\{\scriptsize$\pm21.11$}}
\\
{\scriptsize 13.1\,M\,/\,\textit{13.1}\,M} & Train-from-scratch
& \shortstack{0.823\\{\scriptsize$\pm0.23$}}
& \shortstack{0.353\\{\scriptsize$\pm0.06$}}
& \shortstack{1.46\\{\scriptsize$\pm0.21$}}
& \shortstack{0.127\\{\scriptsize$\pm0.03$}}
& \shortstack{2.69\\{\scriptsize$\pm0.42$}}
& \shortstack{1.76\\{\scriptsize$\pm0.43$}}
\\
& Fine-tuned
& \shortstack{0.803\\{\scriptsize$\pm0.22$}}
& \shortstack{\underline{0.344}\\{\scriptsize$\pm0.05$}}
& \shortstack{\underline{1.35}\\{\scriptsize$\pm0.21$}}
& \shortstack{\underline{0.125}\\{\scriptsize$\pm0.02$}}
& \shortstack{2.67\\{\scriptsize$\pm0.40$}}
& \shortstack{1.73\\{\scriptsize$\pm0.37$}}
\\
\midrule

MonSter & Zero-shot
& \shortstack{11.207\\{\scriptsize$\pm3.79$}}
& \shortstack{25.448\\{\scriptsize$\pm15.28$}}
& \shortstack{65.17\\{\scriptsize$\pm19.47$}}
& \shortstack{15.380\\{\scriptsize$\pm4.83$}}
& \shortstack{78.64\\{\scriptsize$\pm7.96$}}
& \shortstack{76.56\\{\scriptsize$\pm8.02$}}
\\
{\scriptsize 47.8\,M\,/\,\textit{23.0}\,M} & Train-from-scratch
& \shortstack{0.886\\{\scriptsize$\pm0.25$}}
& \shortstack{0.384\\{\scriptsize$\pm0.06$}}
& \shortstack{1.56\\{\scriptsize$\pm0.50$}}
& \shortstack{0.144\\{\scriptsize$\pm0.04$}}
& \shortstack{3.04\\{\scriptsize$\pm0.52$}}
& \shortstack{1.99\\{\scriptsize$\pm0.49$}}
\\
& Fine-tuned
& \shortstack{0.795\\{\scriptsize$\pm0.21$}}
& \shortstack{0.355\\{\scriptsize$\pm0.05$}}
& \shortstack{1.44\\{\scriptsize$\pm0.21$}}
& \shortstack{0.130\\{\scriptsize$\pm0.03$}}
& \shortstack{2.77\\{\scriptsize$\pm0.46$}}
& \shortstack{1.74\\{\scriptsize$\pm0.43$}}
\\
\midrule

DEFOM-Stereo & Zero-shot
& \shortstack{8.712\\{\scriptsize$\pm2.80$}}
& \shortstack{9.020\\{\scriptsize$\pm3.40$}}
& \shortstack{41.63\\{\scriptsize$\pm11.07$}}
& \shortstack{11.248\\{\scriptsize$\pm3.66$}}
& \shortstack{80.31\\{\scriptsize$\pm8.92$}}
& \shortstack{72.48\\{\scriptsize$\pm11.27$}}
\\
{\scriptsize 43.3\,M\,/\,\textit{18.5}\,M} & Train-from-scratch
& \shortstack{0.889\\{\scriptsize$\pm0.24$}}
& \shortstack{0.373\\{\scriptsize$\pm0.06$}}
& \shortstack{1.64\\{\scriptsize$\pm0.50$}}
& \shortstack{0.153\\{\scriptsize$\pm0.04$}}
& \shortstack{3.11\\{\scriptsize$\pm0.55$}}
& \shortstack{1.94\\{\scriptsize$\pm0.45$}}
\\
& Fine-tuned
& \shortstack{0.831\\{\scriptsize$\pm0.23$}}
& \shortstack{0.365\\{\scriptsize$\pm0.05$}}
& \shortstack{1.59\\{\scriptsize$\pm0.21$}}
& \shortstack{0.144\\{\scriptsize$\pm0.04$}}
& \shortstack{2.87\\{\scriptsize$\pm0.50$}}
& \shortstack{1.83\\{\scriptsize$\pm0.44$}}
\\
\midrule

\shortstack[l]{FoundationStereo\\[-1pt]
{\scriptsize 374.5\,M\,/\,\textit{39.2}\,M}}
& \raisebox{2pt}{Zero-shot}
& \shortstack{4.086\\{\scriptsize$\pm1.56$}}
& \shortstack{2.614\\{\scriptsize$\pm0.22$}}
& \shortstack{11.98\\{\scriptsize$\pm2.97$}}
& \shortstack{4.261\\{\scriptsize$\pm1.37$}}
& \shortstack{32.49\\{\scriptsize$\pm36.72$}}
& \shortstack{28.70\\{\scriptsize$\pm34.91$}}
\\

\midrule

HDiT-Stereo \emph{Base} (Ours) & Mean
& \shortstack{0.836\\{\scriptsize$\pm0.22$}}
& \shortstack{0.355\\{\scriptsize$\pm0.05$}}
& \shortstack{1.59\\{\scriptsize$\pm0.19$}}
& \shortstack{0.192\\{\scriptsize$\pm0.05$}}
& \shortstack{3.00\\{\scriptsize$\pm0.57$}}
& \shortstack{1.80\\{\scriptsize$\pm0.34$}}
\\
 & Best-1-of-10
& \shortstack{0.769\\{\scriptsize$\pm0.21$}}
& \shortstack{0.310\\{\scriptsize$\pm0.05$}}
& \shortstack{1.32\\{\scriptsize$\pm0.00$}}
& \shortstack{0.164\\{\scriptsize$\pm0.04$}}
& \shortstack{2.46\\{\scriptsize$\pm0.41$}}
& \shortstack{1.57\\{\scriptsize$\pm0.31$}}
\\
\midrule

HDiT-Stereo \emph{Full} (Ours) & Mean
& \shortstack{\textbf{0.736}\\{\scriptsize$\pm0.21$}}
& \shortstack{\textbf{0.309}\\{\scriptsize$\pm0.05$}}
& \shortstack{\textbf{1.29}\\{\scriptsize$\pm0.12$}}
& \shortstack{0.147\\{\scriptsize$\pm0.04$}}
& \shortstack{\textbf{2.21}\\{\scriptsize$\pm0.37$}}
& \shortstack{\textbf{1.50}\\{\scriptsize$\pm0.33$}}
\\
 & Best-1-of-10
& \shortstack{0.699\\{\scriptsize$\pm0.20$}}
& \shortstack{0.282\\{\scriptsize$\pm0.05$}}
& \shortstack{1.17\\{\scriptsize$\pm0.00$}}
& \shortstack{0.133\\{\scriptsize$\pm0.03$}}
& \shortstack{2.04\\{\scriptsize$\pm0.36$}}
& \shortstack{1.40\\{\scriptsize$\pm0.33$}}
\\

\bottomrule
\end{tabular}%
}
\end{table*}

To quantitatively evaluate our generative stereo model, we draw ten samples per scene and report two statistics: the \emph{mean} of each metric averaged across all samples, and the oracle \emph{best-1-of-10}, which selects the sample with lowest error using ground truth and is therefore unavailable at inference time. We include the latter only to characterize the diversity of the learned distribution, and all practical comparisons use the mean.

In addition to our full model in Fig.~\ref{fig:model_pipeline}, we report a \emph{Base} variant in Table~\ref{tab:main_1000} that is trained only with the disparity objective and removes both the Shared Photometric and Shared Semantic encoders. This isolates the contribution of the UDF prediction head and shared encoder conditioning beyond the underlying generative HDiT. Additional ablations are reported in Appendix~\ref{app:ablation}, with implementation details in Appendix~\ref{app:implementation}.



\label{sec:main_results}

We use metrics that emphasize contour accuracy: \emph{Boundary EPE} is mean absolute disparity error within a $2$-pixel band around ground-truth contours; \emph{Chamfer} \citep{borgefors1988hierarchical} is standard symmetric mean nearest-neighbor contour distance; and \emph{Hausdorff-99} \citep{huttenlocher1993comparing} is a variant of Hausdorff distance that uses the $99$th percentile of pooled bidirectional distances to reduce outlier sensitivity. Alongside these boundary metrics, we also report standard global measures of EPE and Bad-$X$, the percentage of pixels with disparity error above $X$ pixels~\citep{middleburyStereoMetrics}. 

The complete results are shown in Table~\ref{tab:main_1000}. We find that our model is the most accurate on every boundary metric. The improvement is not entirely attributed to an overall increase in model capacity because our \(23.4\)M trainable parameters closely match MonSter’s \(23.0\)M. We find that the baselines' zero-shot metrics are an order of magnitude worse than their performance in other training regimes, suggesting relatively poor zero-shot transfer from SceneFlow to our dataset. We note that our model's improvement near boundaries does not come at the cost of degraded overall disparity accuracy. Our model's Bad-1 and Bad-2 scores also indicate that its residual error is concentrated in fewer pixels.


\subsection{Recovered shape fidelity}
\label{sec:shape_prior}

\begin{figure}[!htbp]
    \centering
    \includegraphics[width=0.9\linewidth]{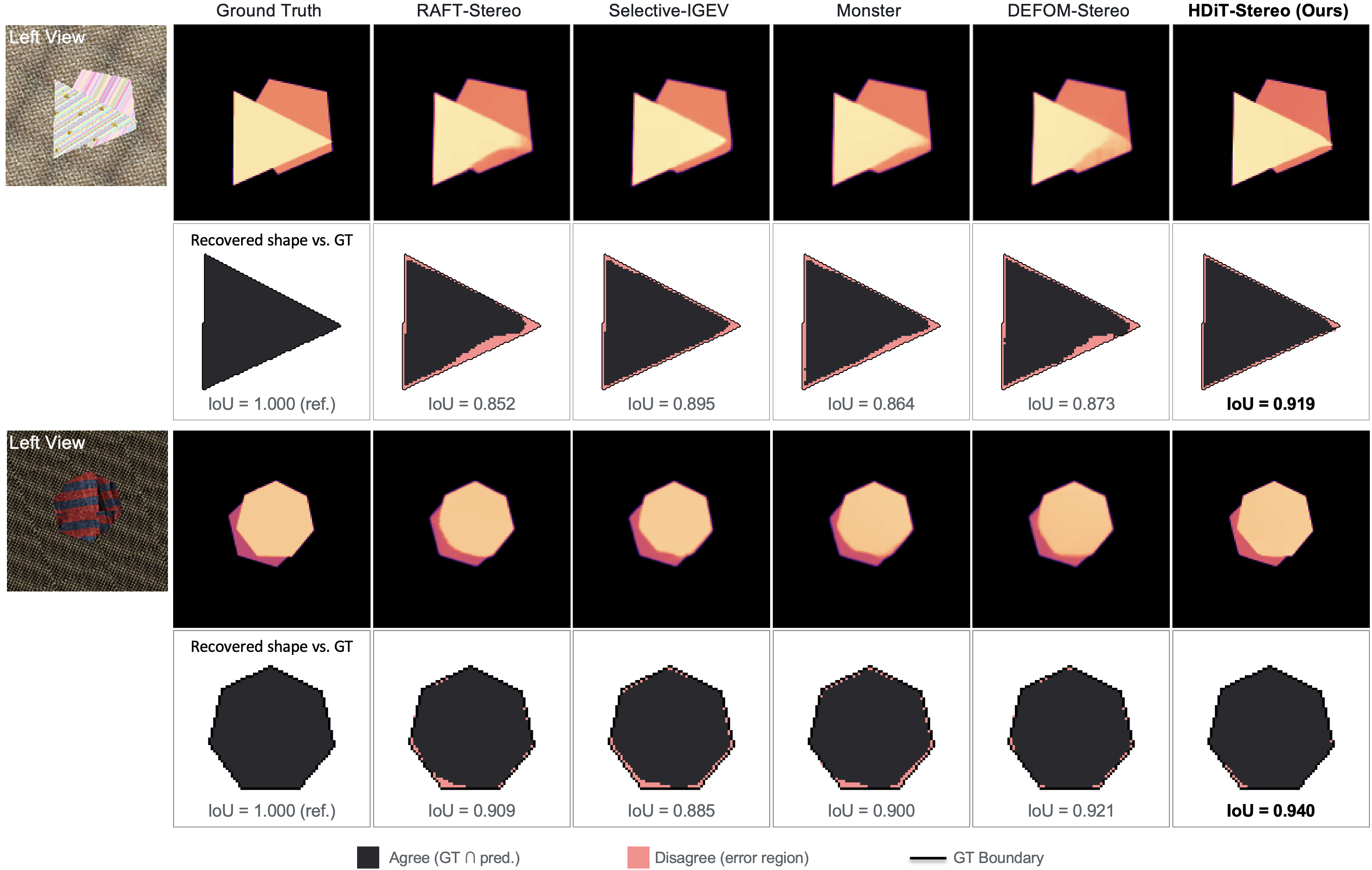}
    \caption{\textbf{Recovered foreground shapes.}
    For each scene, the upper row shows disparity maps with ground truth on the left; the lower row shows recovered foreground shapes overlaid on ground truth. Dark regions indicate agreement, orange regions indicate disagreement, and solid black lines mark ground-truth contours. Our model achieves the highest shape IoU.}
    \label{fig:shape_prior_qualitative}
\end{figure}

Boundary error measures where a contour sits, but it does not measure which object a boundary belongs to (i.e., figure-ground) or how closely it aligns with each object's global shape. We evaluate how closely each model’s recovered contours agree with the known object shapes 
using a test set of two-object scenes. Given a model's predicted disparity map, we segment the object that is nearest (and therefore unoccluded) and evaluate its shape in two ways. First, we measure its agreement ($1{-}\mathrm{IoU}$) with the true object shape as projected into the left image. Second, we find the optimal affine-plus-translation transformation that aligns it to the closest unprojected library shape and compare its pixelwise distances to the boundary with those of the ideal polygon of its ground-truth class. Their Euclidean difference defines \emph{distance to ideal}.


\begin{table}[!htbp]
\centering
\caption{\textbf{Foreground shape accuracy across 100 test scenes.} Lower is better for all metrics. \textbf{Bold} marks the best value and \underline{underline} marks the second-best value among competing methods.}
\label{tab:shape-prior}
\begin{tabular}{lcccc}
\toprule
Method & $n$ & Mean $1{-}\mathrm{IoU}$ $\downarrow$
& Dist.\ to Ideal (mean) $\downarrow$
& Dist.\ to Ideal (median) $\downarrow$ \\
\midrule
RAFT-Stereo              & 100 & 0.0238                    & 1.692                    & 1.568 \\
Selective-IGEV           & 100 & \underline{0.0231}        & \underline{1.684}        & \underline{1.506} \\
DEFOM-Stereo             & 100 & 0.0243                    & 1.733                    & 1.604 \\
MonSter                  & 100 & 0.0245                    & 1.753                    & 1.564 \\
\textbf{HDiT-Stereo (Ours)} & 100 & \textbf{0.0223}           & \textbf{1.631}           & \textbf{1.486} \\
\bottomrule
\end{tabular}
\end{table}

Our model is closest to the floor on all three columns of Table~\ref{tab:shape-prior}. Qualitatively, Figure~\ref{fig:shape_prior_qualitative} shows that all methods recover the correct polygon and the correct coarse disparity, but that their contours differ, with disagreement concentrating along the partially occluded portions of the rear object. 
These results show that HDiT-Stereo yields contours that more closely match the underlying polygons, but they do not by themselves establish that the model is using a stronger global shape prior, because reduced local edge bleeding could produce a similar effect.

\subsection{Uncertainty}
\label{sec:uncertainty}

\begin{figure}[!hbtp]
\centering
\includegraphics[width=0.9\linewidth]{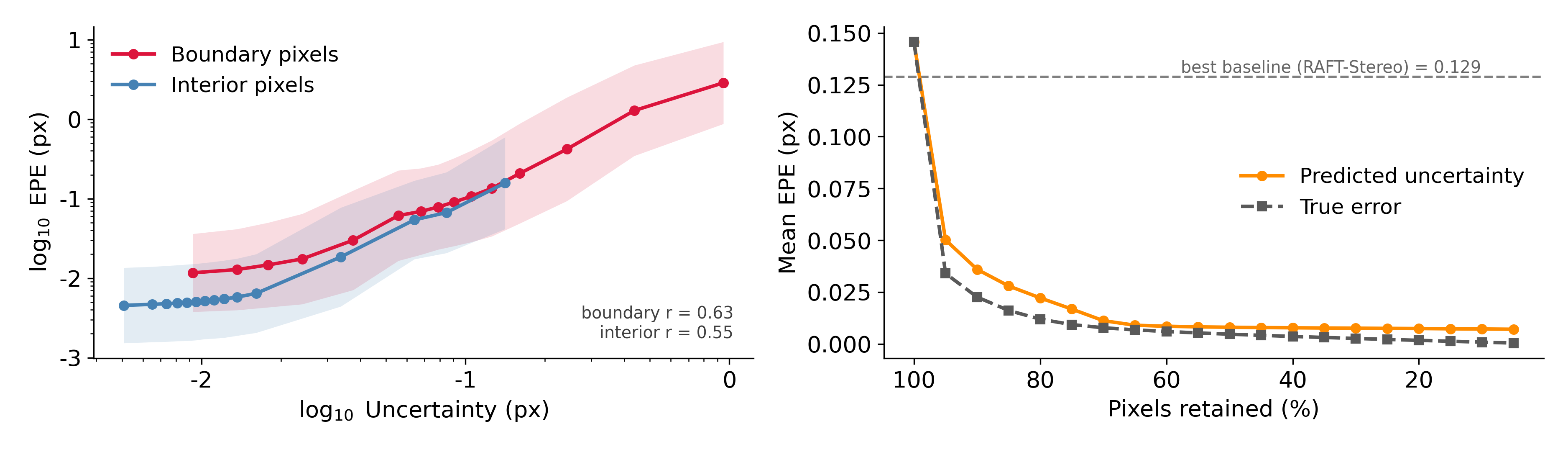}
\caption{\textbf{Sample variation identifies unreliable predictions.}
\textbf{Left:} Mean disparity error versus sample standard deviation, binned separately for boundary and interior pixels across all test scenes. Higher uncertainty indicates larger errors.
\textbf{Right:} Mean error among pixels retained by predicted uncertainty (orange) or ground-truth error (gray). Removing uncertain pixels substantially reduces error without requiring ground truth for selection. Computed on a 100-scene test subset, and the dashed line is RAFT-Stereo on the same subset.}
\label{fig:uncertainty}
\end{figure}

Since our model is generative, we can use multiple samples to produce uncertainty estimates, taking the per-pixel standard deviation of disparity across 10 samples as uncertainty and the per-pixel median as the point prediction, on a random 100-scene subset of the test set. We evaluate whether the per-pixel uncertainty correlates with true disparity error at boundary and interior pixels (Fig.~\ref{fig:uncertainty}, left) and whether it supports effective sparsification by discarding the least-confident pixels and measuring the remaining error (Fig.~\ref{fig:uncertainty}, right).

We find that our model's uncertainty is $5\times$ higher at boundaries than in the interior ($0.161$ vs.\ $0.029$\,px). It correlates with true error (boundary $r{=}0.63$, interior $r{=}0.55$) and so concentrates in locations that are hard to resolve rather than reflecting arbitrary sampling noise. The right of  Fig.~\ref{fig:uncertainty} shows that sparsification using the pixel-wise ranking by uncertainty closely tracks that using an oracle-error ranking, with the most confident half of pixels cutting mean disparity error from $0.146$ to $0.008$\,px against $0.005$\,px for the oracle.
Thus, much of the pruning-based accuracy gains can be obtained from the model's own uncertainty, without access to ground truth.

\subsection{Figural scission, global shape, and bimodality}
\label{sec:bimodality}

The bottom-right example in Fig.~\ref{fig:data_example} shows a challenging case where neither the objects' appearance nor their disparity separation can be used to separate their overlap. Unlike the contour-fidelity experiments in Sec.~\ref{sec:shape_prior}, the local disparity and appearance evidence is entirely insufficient to determine the boundary assignment in these cases, and successful grouping relies critically on nonlocal information about contour continuation and object shape. 

\begin{figure}[!htbp]
    \centering
    \includegraphics[width=0.8\linewidth]{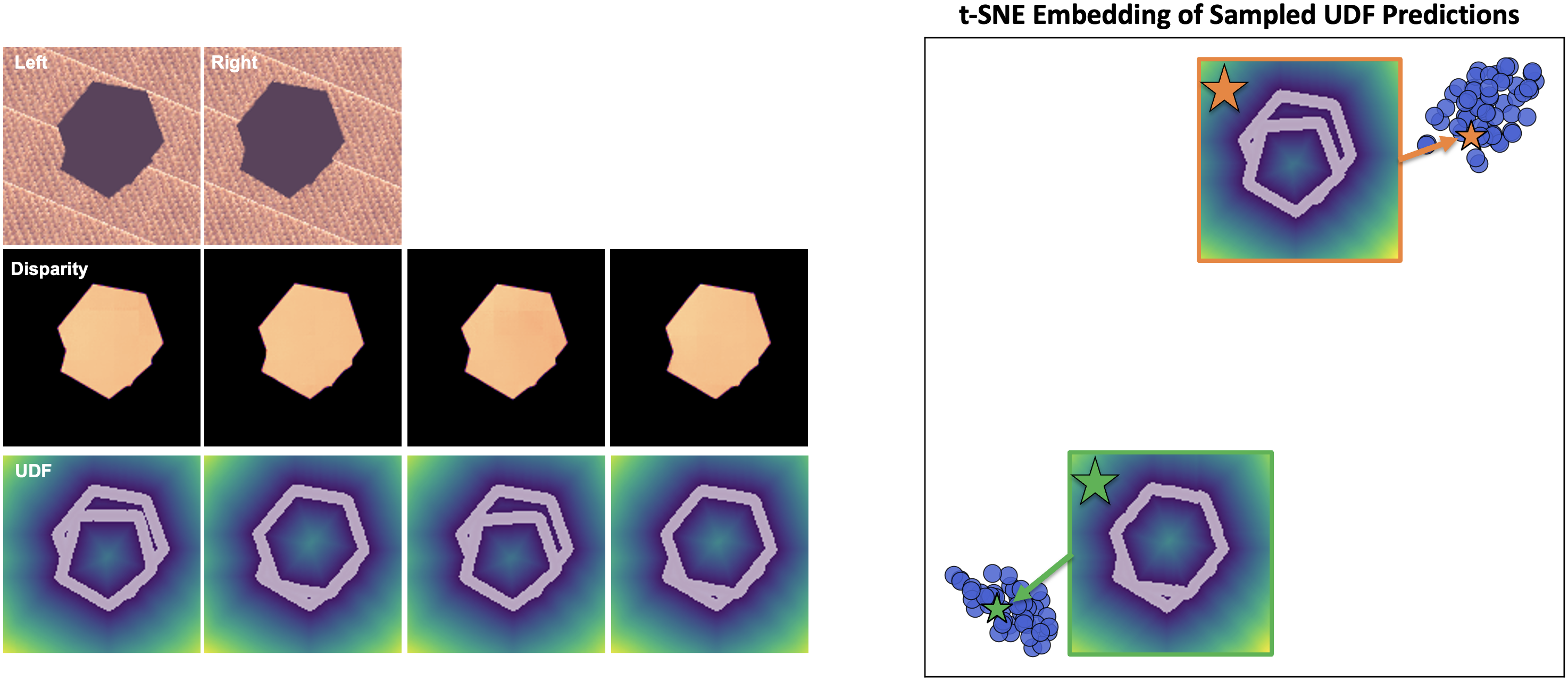}
    \caption{\textbf{Our sampled posterior can be bimodal for ambiguous figure-ground scenes.}
    \textbf{Left:} Four sampled disparity and boundary predictions. Disparity remains consistent while the boundary predictions capture both valid figure-ground interpretations.
    \textbf{Right:} t-SNE embedding of 100 boundary samples, forming two clusters near the two valid contour maps (orange and green stars). Grey bands indicate the predicted zero-level boundary contours.}
    \label{fig:bimodal}
\end{figure}

The figure-ground ordering of the objects is ambiguous in these cases, and humans sometimes perceive them as being bi-stable, with the perceived depth-order flipping spontaneously during sustained viewing~\citep{kellman1991theory,palmer1999vision}. Since our model is generative and produces boundary maps (unsigned distance fields) in addition to disparity, it can mimic this ability by learning to produce a bimodal output distribution over boundary maps. Our model does not always achieve this, but Fig.~\ref{fig:bimodal} shows a successful example. We draw $100$ samples (four shown in figure left) using different noise seeds and embed the predicted distance fields in two dimensions using t-SNE. The disparity samples are all the same as expected, and the boundary samples form two modes that are concentrated near the two true unsigned distance fields. Appendix~\ref{app:bimodal_coverage} provides a quantitative analysis of how often our model successfully recovers both figure-ground interpretations in these types of scenes.


Representing both alternatives naturally requires a multimodal output representation, and sampling from our generative model provides one way to achieve this. A model that emits one answer per pixel would need to choose only one of the boundary maps or an invalid average of the two; and a model that only predicts disparity cannot mimic this effect.

\section{Discussion and limitations}
\label{sec:limitations}
\label{sec:conclusion}

Our controlled experiments show that strong stereo accuracy does not guarantee precise recovery of simple object boundaries. Boundary distortions persist in state-of-the-art disparity models even after in-domain training. HDiT-Stereo improves boundary localization and recovered shape fidelity on this constrained distribution, and its generative boundary representation captures uncertainty and alternative figure–ground interpretations.

These results motivate generative scene modeling as one approach to improving geometric coherence in stereo, but they do not isolate whether the observed disparity improvements arise from stronger global shape reasoning, explicit boundary supervision, or improved local contour prediction. Adding explicit boundary supervision to existing stereo architectures would provide an especially useful comparison.

Our experiments deliberately use a narrow synthetic distribution of regular polygons so that geometric regularities are simple and precisely controlled, and all test scenes are drawn from the same distribution as training. The results therefore establish differences in boundary recovery within this distribution but do not show that they extend to richer object shapes or real-world stereo scenes. Testing more complex and out-of-distribution geometry is an important direction for future work. HDiT-Stereo also requires boundary-field supervision during training, and repeated sampling incurs additional inference cost.


\bibliographystyle{plainnat}
\bibliography{references}

\appendix
\renewcommand{\thesection}{A\arabic{section}}
\renewcommand{\thesubsection}{\thesection.\arabic{subsection}}

\newpage
\section*{Appendix}

\section{Implementation details}
\label{app:implementation}

\subsection{HDiT-Stereo}

\paragraph{Architecture.}
The denoiser is an HDiT with $2\times2$ patches and three levels of depth $[2,2,4]$ and width $[128,256,512]$, using shifted-window attention at the first two levels and global attention at the third. Targets use a fixed $64$\,px normalization scale. The correspondence module searches a $64$\,px disparity range and uses a frozen DINOv2 ViT-S/14 for semantic features. Its soft-argmax disparity receives an auxiliary loss with weight $0.2$.

\paragraph{Encoders.}
The denoiser is conditioned on the stereo pair through three encoders. The \emph{Per View Encoder} processes the left and right views independently with two $3\times3$ convolutional layers (128 channels, ReLU), followed by global average pooling. The two resulting 128-dimensional vectors are concatenated, linearly projected, and added to the noise-level embedding to provide global conditioning to every transformer block.

The \emph{Shared Photometric Encoder} is a three-layer CNN ($3\times3$ convolutions with 32, 64 and 64 channels, ReLU, total stride 2) applied with the same weights to both views. The two sets of L2-normalized features are correlated between views to form a left-view cost volume over disparities from $0$ to $64$ pixels in $2$-pixel increments.

The \emph{Shared Semantic Encoder} extracts frozen DINOv2 ViT-S/14 features from both views, adapts them with a learnable projection, and correlates them between views to form a second, coarser cost volume over the same disparity range. The two cost volumes are concatenated and aggregated into a per-location disparity, which conditions the denoiser through a zero-initialized gate.

\paragraph{Training and Sampling}
We train for $24$ hours on four NVIDIA H200 GPUs with batch size $16$ per device and \texttt{bf16} precision.
We use the EDM loss with soft-min-SNR weighting, $\sigma\in[0.01,160]$, AdamW with a constant learning rate of $5\times10^{-4}$, and an inverse-power EMA.

We sample using EMA weights and DPM++(2M) SDE over $50$ steps with a Karras noise schedule~\citep{karras2022edm}. We draw $10$ samples per scene for quantitative evaluation.

\subsection{Other models}

We use RAFT-Stereo, Selective-IGEV, and DEFOM-Stereo under the MIT license, and MonSter and FoundationStereo in accordance with their non-commercial research restrictions. We train each model for $24$ hours on four NVIDIA H200 GPUs with batch size $16$ per device and \texttt{bf16} precision.  The total compute time for training all of these models is 24 GPU days. Including preliminary work in setting up our training dataset and hyperparameter tuning HDiT-Stereo, the project required approximately 50 GPU days.

\subsection{Inference Computational Cost}

Table~\ref{tab:inference_cost} compares the inference cost of HDiT-Stereo with the regression-based stereo baselines. Because HDiT-Stereo performs iterative diffusion sampling with 50 denoising steps, it requires substantially more computation at inference time than single-forward stereo networks. The Full model requires approximately $3.18$ TFLOPs per sample and $0.318$\,s per image in our evaluation setting, compared with roughly $0.02$--$0.03$\,s for the regression baselines.

This increased inference cost is an important limitation of our approach. Our goal is not to improve inference efficiency, but to study whether generative stereo models can better preserve ambiguous boundary structure and represent multiple plausible solutions. The additional cost reflects the tradeoff of iterative generative inference. Reducing the number of sampling steps, distilling the diffusion model, or adopting faster generative samplers could improve efficiency in future work.

\begin{table}[!htbp]
\centering
\small
\setlength{\tabcolsep}{5pt}
\caption{\textbf{Inference cost} for one $128\times128$ stereo pair on a single A100-40GB in fp32. HDiT-Stereo uses 50-step DPM++(2M) SDE sampling, while the baselines use a single forward pass. FLOPs count multiply-adds as two operations. Runtime is measured per image with a batch of 10, excluding model loading and I/O. Trainable parameters exclude frozen pretrained backbones.}
\label{tab:inference_cost}
\begin{tabular}{lccc}
\toprule
Method
& \shortstack{Params (M)\\ total / trainable}
& \shortstack{GFLOPs\\ per sample}
& \shortstack{Time per\\ sample (s)}
\\
\midrule

Base
& 22.9 / 22.9
& 1,728
& 0.180
\\

Base + Shared Encoders
& 45.4 / 23.4
& 3,182
& 0.318
\\

Base + UDF
& 22.9 / 22.9
& 1,728
& 0.180
\\

\textit{Full} (UDF + Shared Encoders)
& 45.4 / 23.4
& 3,183
& 0.318
\\
\midrule

RAFT-Stereo
& 11.1 / 11.1
& 199
& 0.018
\\

Selective-IGEV
& 13.1 / 13.1
& 218
& 0.025
\\

MonSter
& 47.8 / 23.0
& 207
& 0.027
\\

DEFOM-Stereo
& 43.3 / 18.5
& 207
& 0.018
\\

\bottomrule
\end{tabular}
\end{table}

\subsection{Dataset}

To generate our data, we use the Describable Textures Dataset \cite{cimpoi2014describing} in accordance with their restrictions for research use. All models use the same training dataset of 25,000 images, with scenes generated as described in \ref{sec:data}. 
They are evaluated on 1000 images with the same distribution as the training data.




\section{Ablation of Model Components}
\label{app:ablation}

The ablations in Table~\ref{tab:ablation_1000} isolate the separate contributions of (i) the UDF prediction head and (ii) the two shared encoders, Shared-Photometric and Shared-Semantic. Our \emph{Base} model uses only the disparity objective and removes both shared encoders, and the remaining variants add the UDF head, the shared encoders, or both. The encoders improve both boundary and global metrics, while adding the UDF head further improves boundary and contour accuracy. Combining both components yields the best overall performance, with HDiT-Stereo Full achieving the strongest results across all reported Mean metrics.

\begin{table*}[!htbp]
\centering
\small
\setlength{\tabcolsep}{4pt}
\caption{\textbf{Ablation of HDiT-Stereo components} on the 1000 test scenes ($\pm$ IQR/2 over scenes). \emph{Base} uses only the disparity objective and removes both shared conditioning encoders. UDF denotes the boundary-distance output head; Shared Encoders denotes the pair of Shared Photometric and Shared Semantic encoders, which are added or removed together. Best in \textbf{bold}, second best \underline{underlined} among Mean rows. Best-1-of-10 uses an oracle and is excluded from ranking.}
\label{tab:ablation_1000}

\resizebox{\textwidth}{!}{%
\begin{tabular}{lccclccc|ccc}
\toprule
& & & & & \multicolumn{3}{c|}{\textbf{Boundary \& contour}}
& \multicolumn{3}{c}{\textbf{Global}} \\
\cmidrule(lr){6-8}\cmidrule(lr){9-11}

Variant
& UDF
& \shortstack{Shared\\Encoders}
& Params
& Condition
& Bound.\ EPE $\downarrow$
& Chamfer (px) $\downarrow$
& Hausd.\ 99 (px) $\downarrow$
& EPE $\downarrow$
& Bad1 (\%) $\downarrow$
& Bad2 (\%) $\downarrow$
\\
\midrule

Base
& --
& --
& 22.9\,M
& Mean
& \shortstack{0.836\\{\scriptsize$\pm0.219$}}
& \shortstack{0.355\\{\scriptsize$\pm0.047$}}
& \shortstack{1.59\\{\scriptsize$\pm0.19$}}
& \shortstack{0.192\\{\scriptsize$\pm0.046$}}
& \shortstack{3.00\\{\scriptsize$\pm0.57$}}
& \shortstack{1.80\\{\scriptsize$\pm0.34$}}
\\

&
&
&
&
Best-1-of-10
& \shortstack{0.769\\{\scriptsize$\pm0.21$}}
& \shortstack{0.310\\{\scriptsize$\pm0.05$}}
& \shortstack{1.32\\{\scriptsize$\pm0.00$}}
& \shortstack{0.164\\{\scriptsize$\pm0.04$}}
& \shortstack{2.46\\{\scriptsize$\pm0.41$}}
& \shortstack{1.57\\{\scriptsize$\pm0.31$}}
\\
\midrule

Base + Shared Encoders
& --
& \checkmark
& 45.4\,M
& Mean
& \shortstack{\underline{0.776}\\{\scriptsize$\pm0.209$}}
& \shortstack{\underline{0.335}\\{\scriptsize$\pm0.045$}}
& \shortstack{\underline{1.39}\\{\scriptsize$\pm0.17$}}
& \shortstack{\underline{0.159}\\{\scriptsize$\pm0.037$}}
& \shortstack{\underline{2.30}\\{\scriptsize$\pm0.37$}}
& \shortstack{\underline{1.56}\\{\scriptsize$\pm0.30$}}
\\

&
&
&
&
Best-1-of-10
& \shortstack{0.724\\{\scriptsize$\pm0.20$}}
& \shortstack{0.297\\{\scriptsize$\pm0.05$}}
& \shortstack{1.19\\{\scriptsize$\pm0.00$}}
& \shortstack{0.138\\{\scriptsize$\pm0.03$}}
& \shortstack{2.08\\{\scriptsize$\pm0.35$}}
& \shortstack{1.42\\{\scriptsize$\pm0.31$}}
\\
\midrule

Base + UDF
& \checkmark
& --
& 22.9\,M
& Mean
& \shortstack{0.831\\{\scriptsize$\pm0.230$}}
& \shortstack{0.338\\{\scriptsize$\pm0.046$}}
& \shortstack{1.47\\{\scriptsize$\pm0.19$}}
& \shortstack{0.188\\{\scriptsize$\pm0.049$}}
& \shortstack{3.14\\{\scriptsize$\pm0.67$}}
& \shortstack{1.78\\{\scriptsize$\pm0.37$}}
\\

&
&
&
&
Best-1-of-10
& \shortstack{0.755\\{\scriptsize$\pm0.22$}}
& \shortstack{0.293\\{\scriptsize$\pm0.05$}}
& \shortstack{1.23\\{\scriptsize$\pm0.00$}}
& \shortstack{0.159\\{\scriptsize$\pm0.04$}}
& \shortstack{2.45\\{\scriptsize$\pm0.44$}}
& \shortstack{1.52\\{\scriptsize$\pm0.33$}}
\\
\midrule

\emph{Full} (UDF + Shared Encoders)
& \checkmark
& \checkmark
& 45.4\,M
& Mean
& \shortstack{\textbf{0.736}\\{\scriptsize$\pm0.206$}}
& \shortstack{\textbf{0.309}\\{\scriptsize$\pm0.047$}}
& \shortstack{\textbf{1.29}\\{\scriptsize$\pm0.12$}}
& \shortstack{\textbf{0.147}\\{\scriptsize$\pm0.037$}}
& \shortstack{\textbf{2.21}\\{\scriptsize$\pm0.37$}}
& \shortstack{\textbf{1.50}\\{\scriptsize$\pm0.33$}}
\\

&
&
&
&
Best-1-of-10
& \shortstack{0.699\\{\scriptsize$\pm0.20$}}
& \shortstack{0.282\\{\scriptsize$\pm0.05$}}
& \shortstack{1.17\\{\scriptsize$\pm0.00$}}
& \shortstack{0.133\\{\scriptsize$\pm0.03$}}
& \shortstack{2.04\\{\scriptsize$\pm0.36$}}
& \shortstack{1.40\\{\scriptsize$\pm0.33$}}
\\

\bottomrule
\end{tabular}%
}
\end{table*}

\section{Bimodality for ambiguous two-object scenes}
\label{app:bimodal_coverage}

We complement Sec.~\ref{sec:bimodality} and the successful example in Fig.~\ref{fig:bimodal} by measuring how often HDiT-Stereo captures both plausible figure-ground interpretations for ambiguous two-object scenes. For each ambiguous two-object scene, we draw 50 samples from our model and measure whether both interpretations appear at least once. We quantify the amount of occlusion with the overlap fraction
$$
h = \frac{|M_1 \cap M_2|}{\min(|M_1|, |M_2|)},
$$
where $M_1$ and $M_2$ are the areas of the two objects in the absence of occlusions. This is the fraction of the smaller object's area that overlaps the other object, with $h=0$ when the objects do not touch and $h=1$ when the smaller object lies entirely within the other. We use $h$ instead of IoU because IoU also depends on the objects' size ratio: a small object lying entirely within a larger one has an IoU no higher than their area ratio, even though it is fully covered.

Table~\ref{tab:bimodal_coverage50} reports the percentage of scenes in which both figure-ground interpretations appear, grouped by the overlap fraction $h$ and two values of disparity offset $\Delta d$ between the two objects. Bimodality occurs at best  half of the time, and it trends lower for greater amounts of occlusion. When there is a small non-zero disparity offset ($\Delta d=4$) bimodality is reduced but not eliminated, reflecting that disparity cues remain weak due to the uniform colors of the objects.

\begin{table}[!htbp]
\centering
\caption{Percentage of scenes in which both figure-ground readings appear among 50 samples, grouped by the overlap fraction $h$ and the disparity offset $\Delta d$ between the two objects.}
\label{tab:bimodal_coverage50}
\begin{tabular}{lcccc}
\toprule
Occlusion fraction & 0.2--0.4 & 0.4--0.6 & 0.6--0.8 & 0.8--1.0 \\
\midrule
$\Delta d = 0$ px & 56\% & 31\% & 39\% & 15\% \\
$\Delta d = 4$ px & 31\% & 28\% & 28\% & 13\% \\
\bottomrule
\end{tabular}
\end{table}

\section{Qualitative disparity comparisons}
\begin{figure}[!htbp]
    \centering
    \includegraphics[width=0.95\linewidth]{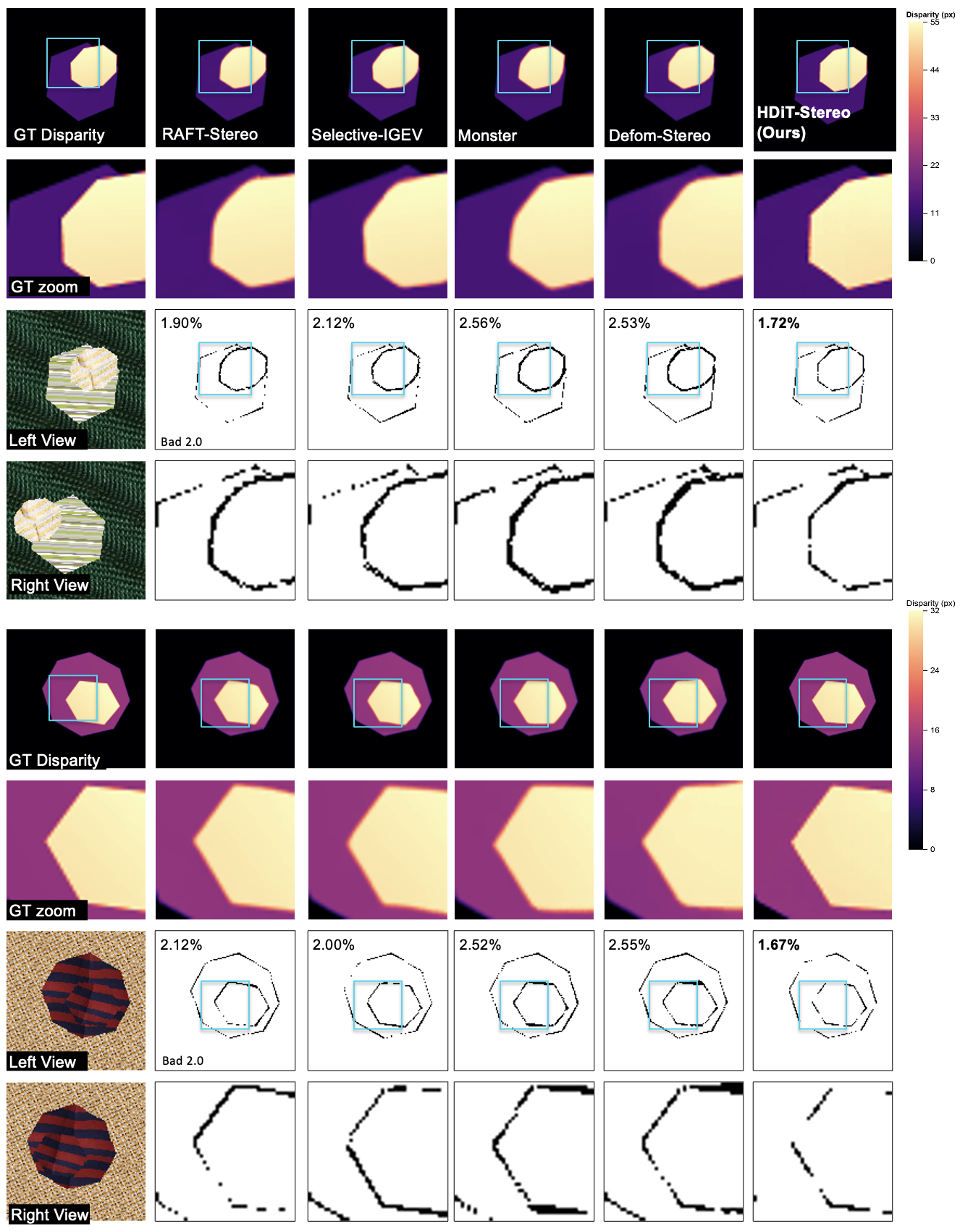}
    \caption{\textbf{Disparity predictions and boundary errors.}
    For each scene, rows show disparity maps, cyan-box enlargements, Bad-2 error maps, and their enlargements; stereo inputs appear at left. Black pixels indicate disparity errors exceeding $2$\,px. HDiT-Stereo better preserves polygonal contours and achieves the lowest Bad-2 rate in both examples.}
    \label{fig:teaser_final}
\end{figure}

Figure~\ref{fig:teaser_final} with different object textures (top four rows) and shared object textures (bottom four rows).

\section{Broader Impact}

This work specifically focuses on evaluating various models on boundaries in synthetic scenes. While we do not expect this specific application could have negative societal impacts, improvements to scene recognition in general could have negative societal effects. In addition, HDiT-Stereo uncertainty predictions should not be used in safety-critical applications, as they are population estimates rather than analytical uncertainty values.

\newpage

\end{document}